\documentclass[10pt,twocolumn]{article}

\usepackage[margin=0.85in]{geometry}
\usepackage{microtype}
\usepackage{times}
\usepackage{setspace}
\usepackage{titlesec}
\usepackage{enumitem}
\usepackage{graphicx}
\usepackage{xcolor}
\usepackage{multicol}
\usepackage{array}
\usepackage{tabularx}
\usepackage{booktabs}
\usepackage{makecell}
\usepackage{ragged2e}
\usepackage{longtable}
\usepackage[sort&compress,numbers]{natbib}
\usepackage{amsmath}
\usepackage[hidelinks,backref=page]{hyperref}
\usepackage{float}
\usepackage{caption}
\usepackage{subcaption}
\usepackage{dblfloatfix}
\usepackage{afterpage}

\hypersetup{
  colorlinks=true,
  linkcolor=black,
  urlcolor=blue,
  citecolor=blue
}
\titleformat{\section}{\large\bfseries}{\thesection.}{0.6em}{}
\titleformat{\subsection}{\normalsize\bfseries}{\thesubsection.}{0.5em}{}
\titleformat{\subsubsection}{\normalsize\itshape}{\thesubsubsection.}{0.5em}{}

\titlespacing*{\section}{0pt}{1.0em}{0.5em}
\titlespacing*{\subsection}{0pt}{0.8em}{0.4em}
\titlespacing*{\subsubsection}{0pt}{0.7em}{0.3em}

\setlist[itemize]{itemsep=2pt, topsep=2pt, leftmargin=1.2em}
\setlist[enumerate]{itemsep=2pt, topsep=2pt, leftmargin=1.2em}

\title{\bf Relationship-Aware Safety Unlearning for Multimodal LLMs\thanks{
\textbf{AI-use disclosure.} Portions of the synthetic data used in preliminary experiments
(images and/or captions) were generated using Google Gemini models. All generated outputs
were reviewed and curated by the authors; the authors take full responsibility for the final
manuscript, experiments, and conclusions.}}

\author{
Vishnu Narayanan Anilkumar \\
Florida International University
\and
Abhijith Sreesylesh Babu \\
Florida International University
\and
Trieu Hai Vo \\
Florida International University
\and
Mohankrishna Kolla \\
Florida International University
\and
Alexander Cuneo \\
Florida International University
}

\date{}

\begin{document}
\RaggedRight
\setstretch{1.05}

\maketitle

\begin{abstract}
Generative multimodal models can exhibit safety failures that are inherently \emph{relational}:
two benign concepts can become unsafe when linked by a specific action or relation
(e.g., \texttt{child}--\texttt{drinking}--\texttt{wine}). Existing unlearning and concept-erasure
approaches often target isolated concepts or image--text pairs, which can cause collateral
damage to benign uses of the same objects and relations. We propose \emph{relationship-aware
safety unlearning}: a framework that explicitly represents unsafe object--relation--object (O--R--O)
tuples and applies targeted parameter-efficient edits (LoRA) to suppress unsafe tuples while
preserving object marginals and safe neighboring relations. We include CLIP-based experiments
and robustness evaluation under paraphrase, contextual, and out-of-distribution image attacks.
\end{abstract}

\vspace{0.2em}


\section{Introduction}

\subsection{Motivation}

Generative AI models—especially multimodal LLMs—have earned remarkable attention recently owing to their ability in content generation. However, they also possess the capability to produce harmful content. This emphasizes the necessity to design robust defense mechanisms to mitigate such risks. While \emph{unlearning} has emerged as a defense strategy to forget unsafe capabilities, existing methods in multimodal content unlearning largely target specific objects or topics (e.g., weapons) rather than \emph{relationships} that become unsafe only within certain contexts. For instance, \emph{a child} and \emph{a bottle of wine}, which are two separate safe entities, would become unsafe when composed as \emph{a child drinking wine}.

In light of such contextual nuances, our objective is to perform unlearning on multimodal LLMs such that the unsafe relationships learned by the model are forgotten. A key area of concern is to ensure that the model's utility is preserved while unlearning such relationships. For example, merely finetuning the model to forget the instance of a child drinking wine might cause it to forget the concepts of ``child,'' ``wine,'' or the ``drinking'' process in other safe scenarios. We aim to minimize this unwanted collateral damage in the generative capabilities of the models while unlearning unsafe relationships, thus promoting ethical use of AI models.

\subsection{Problem Statement}

Given a pretrained multimodal LLM capable of image generation or vision--language reasoning, we aim to design a procedure that:
\begin{itemize}
  \item selectively suppresses unsafe \textbf{object--relation--object} (O--R--O) tuples (e.g., \texttt{child}--\texttt{drinking}--\texttt{wine}),
  \item preserves safe realizations of the same objects or relations (e.g., \texttt{adult}--\texttt{drinking}--\texttt{wine}; \texttt{child} next to \texttt{wine bottle} with no consumption),
  \item remains robust against prompt obfuscation and compositional paraphrases.
\end{itemize}

\subsection{Objectives}

\begin{enumerate}[leftmargin=2.5em]
  \item \textbf{Relational Safety Taxonomy:} Formalize a schema for unsafe relational tuples with context (actors, actions, objects, attributes, spatial/temporal cues).
  \item \textbf{Relationship-Aware Unlearning:} Develop fine-tuning or parameter-editing procedures (e.g., contrastive unlearning, LoRA masking, causal trace edits) targeted at O--R--O tuples.
  \item \textbf{Context Preservation:} Introduce counterfactual preservation losses and safe exemplars to retain utility on allowed and safe context examples.
  \item \textbf{Red-Team \& Robustness:} Evaluate resistance to prompt fuzzing, synonym swaps, and compositional adversaries.
  \item \textbf{Auditable Safety:} Provide testable acceptance criteria and unit tests for safety regression.
\end{enumerate}

\section{Related Work}
\label{sec:related-work}

Machine unlearning (MU) has been studied in both classical models and modern foundation models. The problem was first framed as removing the influence of specific training samples while keeping overall performance essentially unchanged~\citep{cao2015unlearning, bourtoule2021machine}. Follow-up work looks at MU from systems, theory, and application angles~\citep{nguyen2022survey, xu2023machine, zhang2023review}. More recent surveys and SoK-style papers focus on deep and foundation models, and especially on large language models (LLMs), offering taxonomies, metrics, and threat models~\citep{liu2024survey, liu2025survey, yao2023largellmunlearning, liu2024rethinking, ren2025sokllm}. A common conclusion is that many current methods behave more like ``suppression'' than true removal, and that evaluation practices are still immature.

\vspace{1.0em}

For LLMs, most methods adapt a pre-trained model with gradient-based updates on a \emph{forget set} while trying to preserve performance on a \emph{retain set}. Early work on knowledge unlearning shows that simple gradient-ascent updates on targeted sequences can meaningfully reduce memorization of private facts while keeping language modeling quality~\citep{jang2023knowledge}. \citet{yao2023largellmunlearning} and follow-ups then systematize gradient-ascent style objectives and compare them to retraining baselines. Benchmarks such as TOFU~\citep{maini2024tofu} and R-TOFU for reasoning models~\citep{yoon2025rtofu} formalize forget/retain metrics, while MUSE proposes a six-way evaluation protocol that looks at memorization, privacy leakage, utility, scalability, and repeated unlearning requests~\citep{shi2025muse}.

\vspace{1.0em}

Several methods recast unlearning as a preference optimization problem. Negative Preference Optimization (NPO) improves the stability of naive gradient ascent and reaches a better forget/retain trade-off on TOFU~\citep{zhang2024negativepref}, and SimNPO simplifies the objective and reference-model design while keeping competitive performance~\citep{fan2024simnpo}. Second-order approaches such as SOUL use curvature information to make unlearning updates more stable and efficient~\citep{jia2024soul}. Alongside objective design, there is work on scalable MU procedures in learned databases~\citep{kurmanji2024experimental} and on formal guarantees based on algorithmic stability and differential privacy~\citep{ullah2021stability}. Recent analyses argue that, even with these techniques, many models still retain traces of the supposedly forgotten information and call for more careful evaluation~\citep{liu2024rethinking}.

\vspace{1.0em}

In vision, unlearning has been explored mainly in text-to-image diffusion models. Early methods remove single risky concepts such as nudity, specific characters, or styles by weakening cross-attention paths, steering text embeddings, or reweighting attention, so that prompts containing a flagged token no longer produce the undesired content~\citep{gandikota2023erasing, park2024duo, lyu2024onedadapter}. These techniques show that concepts can be erased at the parameter level rather than by post hoc filters, but they mostly treat each concept in isolation and can unintentionally damage nearby concepts. Later work observes that many safety issues involve combinations of otherwise benign ideas, such as child + alcohol. Concept-combination erasure explicitly encodes such themes and aims to suppress the combinations while preserving the individual concepts, using logic-guided prompt enumeration and feature-space decoupling~\citep{yao2025erasingcombination}. FADE goes further by adding adjacency-aware losses that jointly encourage target erasure, preservation of semantically related neighbors, and consistency with the original guidance process, giving more precise control over what is forgotten and what is retained~\citep{thakral2025fade}. Multi-concept erasure and survey work on concept suppression help organize this growing space~\citep{zhao2024separableerasure, bian2025conceptsuppressionsurvey}.

\vspace{1.0em}

Multimodal encoders, such as CLIP-style models, raise a complementary set of concerns. These encoders align images and text for retrieval, reranking, and moderation, and often sit upstream of safety or policy filters, so removing unsafe associations without degrading overall utility is important. MultiDelete performs MU at the level of image--text pairs, decoupling specified pairs while preserving unimodal quality and overall alignment on retained data~\citep{cheng2024multidelete}. CLIPErase proposes a three-part objective for CLIP-like encoders: a forgetting term that reduces similarity on targeted pairs, a retention term that reinforces alignment on safe data, and a consistency term that keeps embeddings close to the base model so that zero-shot performance is not heavily degraded~\citep{yang2025cliperase}. Other work looks at structure inside the model. Selective pruning methods identify neurons that are especially important for a targeted behaviour and remove them to unlearn specific capabilities~\citep{pochinkov2024selectivepruning}, while LLM-Eraser extends this idea to large language models with explicit localization and pruning stages~\citep{zhang2025llmeraser}. In multimodal LLMs, modality-aware pruning allocates different pruning budgets to text and vision pathways to selectively weaken unsafe associations while preserving core skills~\citep{liu2025modalityprune}.

\vspace{1.0em}

Across diffusion and multimodal encoders, most existing work either removes single objects or concepts, or breaks specific image--text \emph{pairs}. Many safety-sensitive behaviours, however, are relational: they depend on how two objects interact under a relation (for example, \emph{child--drinking--beer}) rather than on the objects alone. Generator-side concept-combination erasure~\citep{yao2025erasingcombination} and adjacency-aware erasure~\citep{thakral2025fade} move in this direction on the synthesis side by suppressing unsafe combinations while preserving marginals and neighbors. At the encoder level, current multimodal MU methods still operate on samples, pairs, or global concepts and do not explicitly represent relational structure between objects and relations. Relation-level unlearning in contrastive encoders and large multimodal models therefore remains largely unexplored.

\vspace{1.0em}

Our work targets this gap with \emph{relation-aware unlearning}. We aim to forget unsafe triplets \(O_1\!-\!R\!-\!O_2\) while preserving the marginal concepts \(O_1\) and \(O_2\) and keeping safe uses of \(R\) with other objects intact. We build a small relation graph that marks which edges (relations) are unsafe and which adjacent relations should be preserved, and we train lightweight adapters with a multi-objective loss that encourages strong forgetting on unsafe relations, retention of the participating objects, preservation of benign neighboring relations, and geometric consistency with the base encoder. This complements generator-side unlearning in diffusion models and encoder-side multimodal MU, and is directly useful for relation-sensitive tasks such as zero-shot classification, bidirectional retrieval, and content moderation.

\section{Methodology}

Our method introduces a systematic framework for unlearning unsafe relationships in multimodal large language models while preserving the safe concepts that are necessary for the utility of the model. Our approach consists of two components: construction of a relational graph that explicitly represents the object--relation structures related to the unsafe relationship, and a targeted parameter-editing procedure to selectively weaken the model's representations of the unsafe relations while preserving the model's performance on other concepts. Together these components form a pipeline that mitigates the selected unsafe relations from the model's knowledge space.

\begin{figure*}[!t]
    \centering
    \includegraphics[width=0.96\textwidth]{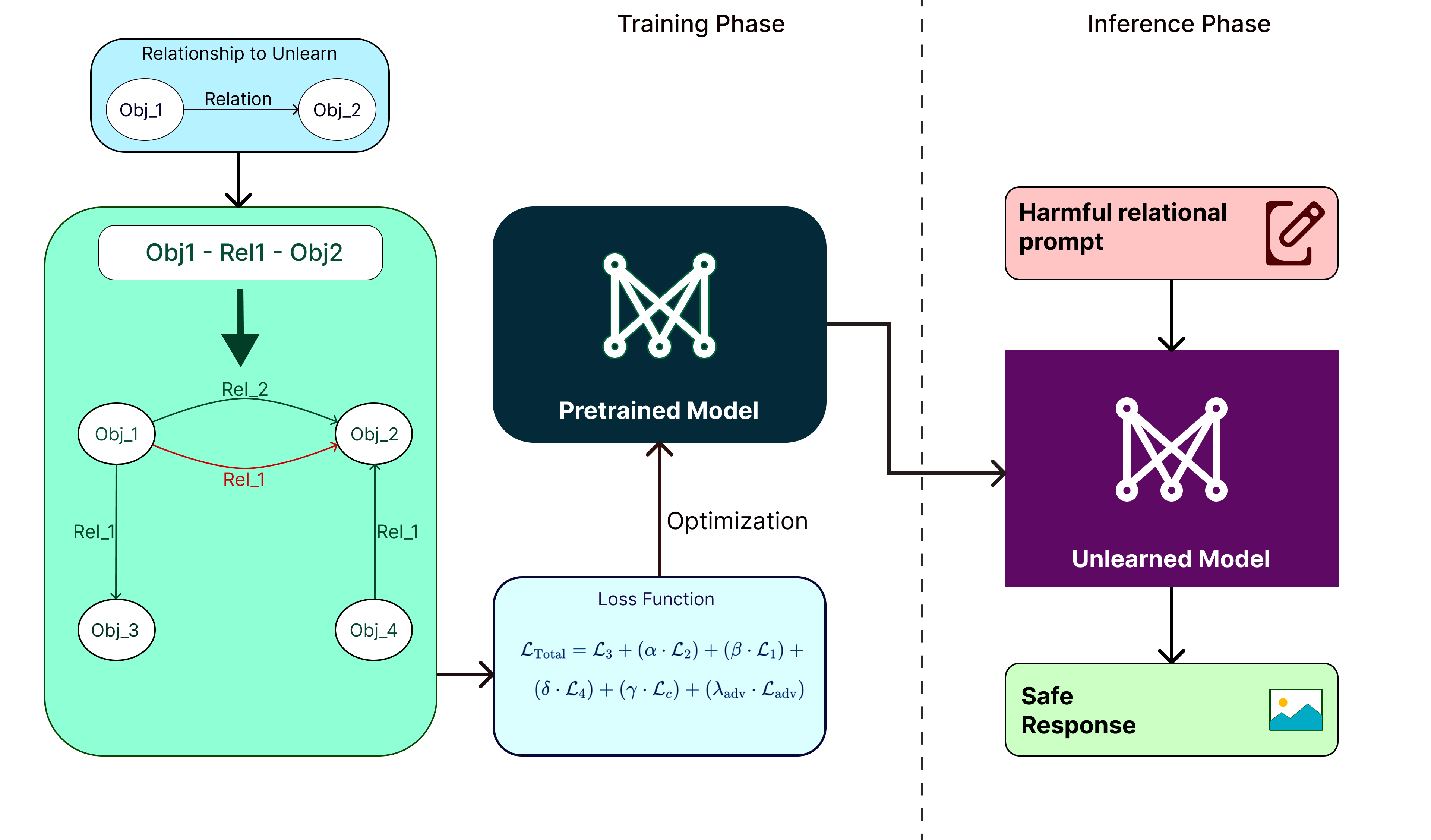}
    \caption{Overall architecture}
    \label{fig:arch}
\end{figure*}

\subsection{Graph-based Representation}

Our first objective is to define a structured graph that captures unsafe relationships. The relationship graph is a graph data structure where each node represents the objects involved in the relationships and each edge represents a kind of relationship. By representing the target relationship as a graph, we are able to achieve a formal segregation of what to unlearn and what to preserve. In case of unlearning a relationship \(O_1-R_1-O_2\), we form a graph where \(O_1\) and \(O_2\) are nodes connected by an edge representing \(R_1\). In order to preserve \(R_1\) in other object combinations, we add new objects \(O_3\) and \(O_4\), with edges representing \(R_1\) connecting \(O_1\) and \(O_2\) to \(O_3\) and \(O_4\), respectively. By doing this, we are able to preserve objects other than \(O_1\) and \(O_2\), and at the same time preserve the relationship \(R_1\) between two other objects. Now we need to preserve the relationship between \(O_1\) and \(O_2\) in some other relation. For that, we add an edge \(R_2\) between \(O_1\) and \(O_2\) to make sure that the objects \(O_1\) and \(O_2\) are preserved in combination if their relationship is not the same as the targeted harmful relation.

\subsection{Parameter Editing}

The graph formed from these object--relationship schemas serves as the reference for the unlearning optimization function. Given a graph for the relation \(O_1-R_1-O_2\), we need to unlearn the edge \(R_1\) between \(O_1\) and \(O_2\). To calculate the loss for this relationship, we encode the text-based caption and corresponding images of this relation to an embedding space. For this, we can use the encoder of the targeted model. We then take the cosine similarity between the embeddings as a loss metric for the relation. In addition to this, we need to preserve the rest of the nodes and edges in the graph.

In the unlearning stage, we finetune the target model by employing a low-rank adapter~\cite{hu2022lora} on any of its attention and projection layers. The key advantage of having an adapter is that it decouples the model from its unlearned version while helping reduce collateral damage to adjacent classes.

\vspace{1.0em}

The loss function for training the model consists of various components for unlearning unsafe relations, preserving safe relations, and mitigating potential adversarial attacks from users. Loss term \(\mathcal{L}_1\) pulls the concerned edges in the graph that represent safe contexts. \(\mathcal{L}_2\) pulls the concerned nodes in the graph that represent safe objects. \(\mathcal{L}_3\) pushes the edge that represents the unsafe relationship that we want to unlearn. \(\mathcal{L}_4\) pulls the model's performance on random concepts unrelated to the unlearned concept. \(\mathcal{L}_c\) ensures consistency by making sure the unlearned model works similarly to the original model. \(\mathcal{L}_{\text{adv}}\) pushes the adversarial variants of texts to mitigate possible attacks. All the loss terms work together to form the loss function used in optimization:

\begin{equation*}
\mathcal{L}_{\text{total}} = \mathcal{L}_{3} + \alpha \mathcal{L}_2 + \beta \mathcal{L}_1 + \delta \mathcal{L}_4 + \gamma \mathcal{L}_c + \lambda_{\text{adv}} \mathcal{L}_{\text{adv}}.
\end{equation*}

\section{Experiment Results and Analysis}

To experiment with our method, we removed a safe relationship from the CLIP model~\cite{radford2021learning}. This allowed us to ensure that the removal was a result of our method and not of the model's pre-training phase. Our experiment aimed to remove the relationship of a kid eating a hamburger.

\subsection{Metrics}

The CLIP model maps text and images into a shared embedding space, and the semantic alignment between them is measured using cosine similarity:
\[
\text{Cosine Similarity} = \cos(\theta) = \frac{\mathbf{t} \cdot \mathbf{i}}{|\mathbf{t}| |\mathbf{i}|}
\]
where \(\mathbf{t}\) and \(\mathbf{i}\) are text and image embeddings, respectively. Higher cosine similarity indicates stronger semantic association. Within our selective learning framework, the goal is to reduce cosine similarity for unsafe relationships while maintaining high similarity for safe and neutral nodes and edges. We evaluate model performance by comparing cosine similarity scores before and after learning to quantify forgetting and utility preservation.

\subsection{Datasets}

The data that is used for finetuning consists of image--text pairs for each of the terms in the loss function. The images in the dataset are synthetically generated using the vision--language model \textit{gemini-2.0-flash-preview-image-generation}. This ensures that the data used for finetuning contains necessary features that are useful in the optimization process. The corresponding texts for the images are generated using the model \textit{gemini-2.5-flash}. In this way, we can generate datasets for unlearning any relationships even if explicit datasets are not available.

\subsection{Resource Requirement}

The experiments in this project required significant computational resources due to the size of the CLIP model and the repeated forward passes necessary for the LoRA-based unlearning process. All training and evaluation were performed on a Ruby server equipped with an NVIDIA A100 GPU (40GB RAM). This provided the necessary memory capacity to handle large batch operations, embedding multidimensional data, and multiple loss components simultaneously. The 40GB GPU memory was particularly important because both text and image encoders were active during training, and LoRA updates introduced additional learnable parameters. This hardware configuration ensured stable training, prevented out-of-memory errors, and allowed the knowledge removal process to be completed efficiently within a reasonable timeframe.

\subsection{Baseline Methods}

To evaluate the effectiveness of our method, we compare our method with a baseline unlearning procedure. In the baseline setting, we only use the term \(\mathcal{L}_3\) as a loss function to finetune the model. This method pushes the target unsafe relationship but fails to preserve safe nodes and edges. In contrast, our full method uses a fully optimized loss function designed to preserve strong performance in safe concepts and relations.

\subsection{Training Setup}

The model was trained using the AdamW optimizer with a learning rate of \(1\times10^{-3}\), a batch size of 32, and over 3 epochs. The training process was monitored through a loss function convergence graph. It illustrates the evolution of each loss component over iterations. The graph confirms that Unsafe Push loss (\(\mathcal{L}_3\)) and Adversarial Push loss (\(\mathcal{L}_{\text{adv}}\)) decrease progressively, effectively promoting the forgetting of unsafe information. Meanwhile, Consistency loss (\(\mathcal{L}_c\)) remains close to zero, ensuring minimal change in safe and neutral representations. At the same time, Pull loss (\(\mathcal{L}_{\text{pull}} = \mathcal{L}_1 + \mathcal{L}_2 + \mathcal{L}_4\)) remains stable, preserving important relationships between safe and neutral nodes. Overall, the loss curves show a well-balanced training process, achieving selective forgetting without compromising the model's usefulness.

\begin{figure}[t]
    \centering
    \includegraphics[width=\linewidth]{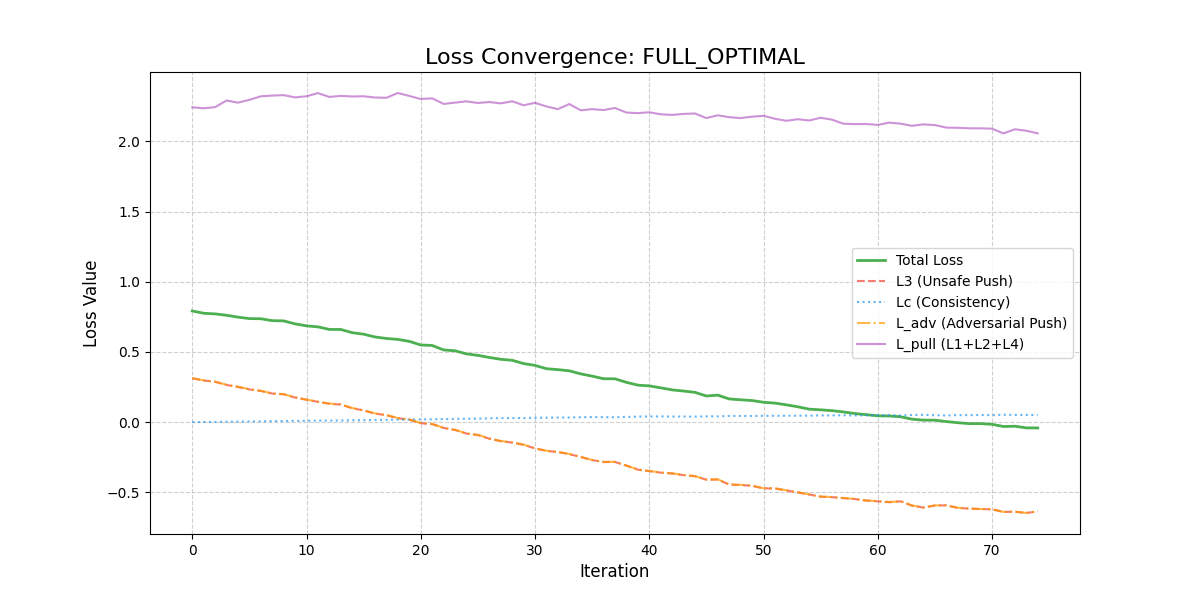}
    \caption{Loss Convergence}
    \label{fig:loss}
\end{figure}

\subsection{Results \& Discussion}

Experimental evaluations were conducted on three different scenarios—easy, medium, and hard. This allowed for a comprehensive assessment of the model's performance in selectively removing unnecessary information. Three types of attacks were used: Paraphrase Attack, Contextual Attack, and Image Attack. The results in Table~\ref{tab:cosine_comparison} show a significant reduction in cosine similarity after applying the FULL\_OPTIMAL model. The \(\Delta\)cos values were 0.6878, 0.4881, and 0.7012 for paraphrase, contextual, and out-of-distribution (OOD) image attacks, respectively. These results indicate that the model effectively forgets unsafe information across diverse linguistic and visual perturbations. These include rephrased text prompts, contextually embedded phrases, and stylized images. Examples of attacks include a paraphrased prompt ``Youngster taking a bite of a meat patty,'' a contextual scenario in a ``futuristic city street at night,'' and an OOD image resembling Van Gogh's painting style.

Simultaneously, utility preservation was rigorously evaluated to ensure that safe and neutral knowledge remains intact. Table~\ref{tab:preservation_cosine} shows that the absolute drift (\(|\Delta\text{cos}|\)) across cases such as Single Node Preservation, New Safe Edge, New Safe Node, and New Neutral Edge remains minimal, ranging from 0.0115 to 0.0608. Sample cases include images of a happy child, a delicious hamburger, people drinking coffee, or a stunning Eiffel Tower. This confirms that the FULL\_OPTIMAL model achieves a strong balance: it effectively forgets unsafe concepts while preserving safe and neutral relationships, ensuring both selective unlearning and knowledge retention.

\begin{table}[t]
\centering
\small
\caption{Comparison of Cosine Similarity Before and After Optimization}
\label{tab:cosine_comparison}
\begin{tabular}{|l|c|c|c|}
\hline
\textbf{Attack Type} & \textbf{Base Cosine} & \textbf{Optimal Cosine} & \textbf{\(\Delta\)cos (Forgetting)} \\
\hline
Paraphrase & 0.3010 & -0.3868 & 0.6878 \\
\hline
Contextual & 0.2522 & -0.2359 & 0.4881 \\
\hline
OOD Image & 0.2716 & -0.4296 & 0.7012 \\
\hline
\end{tabular}
\end{table}

\begin{table}[t]
\centering
\scriptsize
\setlength{\tabcolsep}{3pt}
\renewcommand{\arraystretch}{1.15}
\caption{Comparison of Cosine Similarity for Different Preservation Cases}
\label{tab:preservation_cosine}
\resizebox{\columnwidth}{!}{%
\begin{tabular}{|l|c|c|c|}
\hline
\textbf{Preservation Case} & \textbf{Base Cosine} & \textbf{Optimal Cosine} & \textbf{Abs.\ Drift (\(|\Delta\text{cos}|\))} \\
\hline
Single Node Preservation & 0.2820 & 0.3428 & 0.0608 \\
\hline
New Safe Edge & 0.2158 & 0.2607 & 0.0450 \\
\hline
New Safe Node & 0.2866 & 0.2981 & 0.0115 \\
\hline
New Neutral Edge & 0.3224 & 0.2754 & 0.0470 \\
\hline
\end{tabular}}
\end{table}

\subsection{Ablation Study}

An ablation study was performed to examine the contribution of each individual loss component in the FULL\_OPTIMAL model and to confirm the importance of a balanced objective function for efficient selective unlearning. As illustrated in the comparison graphs (``Ablation Visualization'' and ``Utility Preservation''), the baseline model—relying solely on the core \(\mathcal{L}_3\) unsafe push loss—achieves strong forgetting but suffers from significant utility degradation, exhibiting the largest parameter drift. In contrast, the FULL\_OPTIMAL model, integrating Push, Pull, Consistency, and Adversarial objectives, achieves superior balance. It maximizes the efficiency of forgetting unsafe information while minimizing drift in safe and core relational knowledge, affirming the overall utility preservation capability of the model.

Analysis of specific ablation scenarios further highlights the crucial role of auxiliary loss components. Consistency loss (\(\mathcal{L}_c\)) is important in limiting the overall embedding-space drift. Models lacking \(\mathcal{L}_c\) show significantly higher drift in the fundamental concepts (Node Preservation). Similarly, adversarial loss (\(\mathcal{L}_{\text{adv}}\)) plays a vital role in enhancing generalized forgetting. It ensures that the unlearning effect is reliably extended to paraphrased or slightly altered unsafe prompts. Overall, these findings validate the FULL\_OPTIMAL loss design. It demonstrates that each component is indispensable for achieving both highly accurate forgetting and essential knowledge retention within the CLIP unlearning framework.

\begin{figure*}[!t]
    \centering
    \includegraphics[width=0.96\textwidth,height=0.32\textheight,keepaspectratio]{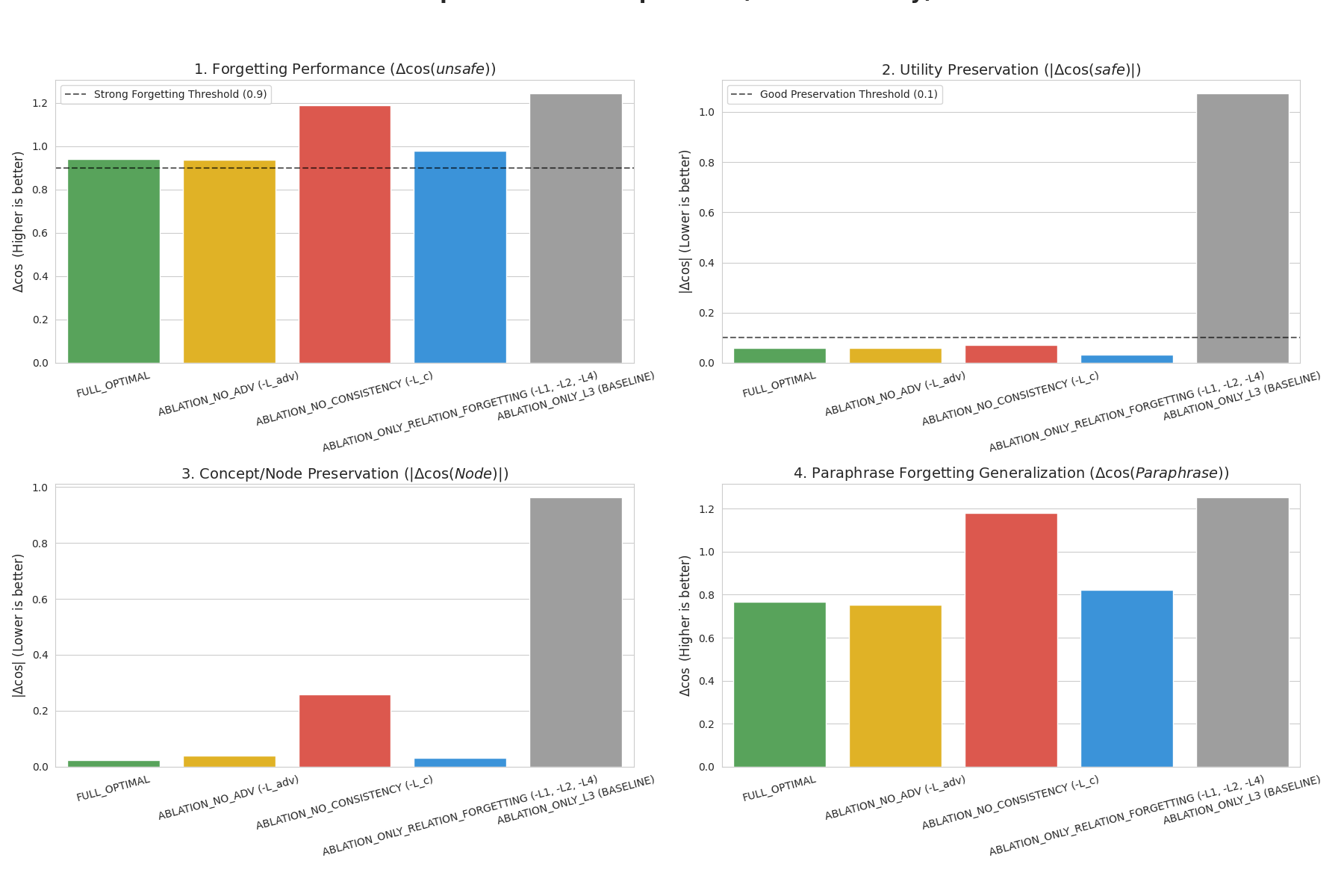}
    \caption{Ablation Visualization}
    \label{fig:ablation}
\end{figure*}

\afterpage{%
\begin{figure*}[!t]
    \centering
    \includegraphics[width=0.96\textwidth,height=0.32\textheight,keepaspectratio]{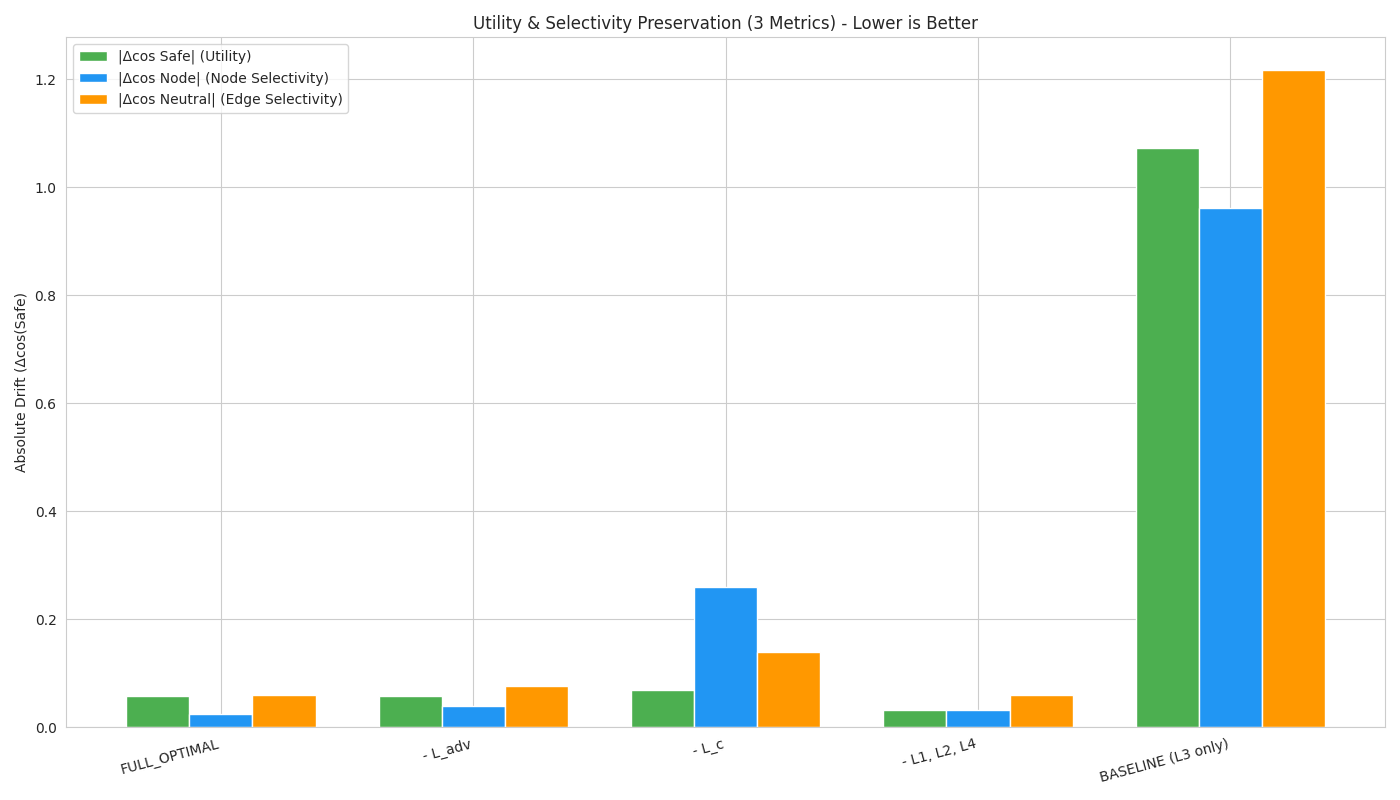}
    \caption{Utility Preservation}
    \label{fig:utility}
\end{figure*}
}

\section{Future works}

The current research establishes a novel framework for relation-aware unlearning by targeting specific O--R--O tuples in multimodal LLMs. While successful, this work opens several exciting avenues for future exploration to enhance the methodology's efficacy, scalability, and applicability.

\subsection{Extending Relational Unlearning Mechanism}

Our current approach utilizes a multi-objective loss function to guide unlearning via LoRA adapters. Future work should investigate more granular and efficient parameter-editing techniques:

\begin{itemize}
    \item \textbf{Causal Tracing and Neuron Pruning:} While mentioned in our objectives, a concrete investigation is needed to identify the minimal set of neurons or attention heads in the multimodal encoder that are responsible for the unsafe O--R--O association. By combining causal tracing with selective pruning, we could achieve \emph{surgical} unlearning, leading to faster updates and potentially zero collateral damage compared to broader fine-tuning methods.
    
    \item \textbf{Hierarchical Graph Structures:} The current relationship graph is relatively flat. We propose extending the graph to incorporate semantic hierarchies (e.g., \texttt{wine} \(\rightarrow\) \texttt{alcohol} \(\rightarrow\) \texttt{beverage}) or attribute-level constraints (e.g., \texttt{child} with the attribute \texttt{under 18}). Unlearning could then propagate through the hierarchy, ensuring that if \texttt{child}--\texttt{drinking}--\texttt{wine} is forgotten, semantically subordinate tuples like \texttt{toddler}--\texttt{drinking}--\texttt{beer} are also suppressed without explicitly labeling every permutation.
\end{itemize}

\subsection{Improving Robustness and Adversarial Mitigation}

The current work includes an adversarial loss term, \(\mathcal{L}_{\text{adv}}\), to push against prompt obfuscation. This area can be significantly deepened:

\begin{itemize}
    \item \textbf{Proactive Adversarial Data Generation:} Instead of relying on simple text paraphrases, future iterations can employ a \emph{Red-Team Generative Model} to actively create the most challenging adversarial prompts (e.g., highly compositional sentences, non-verbal cues in the image, or misspellings). This can be used to continuously refine the \(\mathcal{L}_{\text{adv}}\) term and improve model resistance.
    
    \item \textbf{Compositional Robustness Benchmarking:} We aim to develop a dedicated benchmark that specifically tests compositional generalization \emph{after} unlearning. For example, if the tuple \texttt{child}--\texttt{drinking}--\texttt{wine} is forgotten, the benchmark must verify whether the model retains the ability to generate safe, but compositionally similar scenes, such as \texttt{child}--\texttt{holding}--\texttt{wine glass}. This moves beyond current cosine similarity metrics to evaluate generative reasoning more holistically.
\end{itemize}

\subsection{Scaling and Generalization}

Our initial experiments focused on the CLIP model, a static encoder. Future work needs to focus on scaling this approach to much larger generative multimodal LLMs:

\begin{itemize}
    \item \textbf{Scaling to Generative Models:} We plan to apply relation-aware unlearning directly to the U-Net or cross-attention layers of text-to-image diffusion models. This would require adapting the loss function to minimize the probability of generating the unsafe composition while maximizing the probability of generating safe counterfactuals from the same prompt (e.g., blurring the prohibited object or modifying the subject).
    
    \item \textbf{Parameter-Efficient Scaling:} Given the resource constraints noted in our experiments, we will investigate alternative parameter-efficient fine-tuning (PEFT) methods beyond LoRA, such as DoRA (Weight-Decomposed Low-Rank Adaptation) or QLoRA. These methods may offer better forget/retain trade-offs or faster convergence when unlearning on massive models with limited GPU memory.
\end{itemize}

\subsection{Auditable Safety and Evaluation}

The final step is to formalize the evaluation of successful, relational unlearning into industry-standard metrics:

\begin{itemize}
    \item \textbf{Auditable Acceptance Criteria:} We intend to formalize the acceptance criteria into a quantitative metric, potentially a \emph{Relational Safety Score} (RSS). This score would combine: (1) the suppression rate of unsafe O--R--O tuples, (2) the preservation rate of safe marginal objects and relations, and (3) the model's resistance to red-teaming attacks.
    
    \item \textbf{Downstream Utility Impact:} Beyond the cosine similarity metric, we will rigorously evaluate the impact of unlearning on critical downstream tasks for multimodal encoders, such as zero-shot image classification and bidirectional retrieval on unrelated, safe datasets. This ensures that the utility preservation terms (\(\mathcal{L}_c\) and \(\mathcal{L}_4\)) remain effective in diverse deployment scenarios.
\end{itemize}

\section{Conclusion}

The development and implementation of our relationship-aware safety unlearning system for multimodal LLMs gave us critical insights about the complexity of selective knowledge removal. First of all, our experience showed that effective unlearning requires a well-balanced multi-objective loss function instead of simplistic mechanisms used in related work. Our first baseline model, which used only unsafe push loss to contrast embeddings that were harmful, showed convincingly that forgetting targeted content is far from enough in practice. Only the fully optimized model, featuring safe pull loss, node pull loss, neutral pull loss, consistency loss, and adversarial loss, made it possible to sustain high model utility while yielding strong selective forgetting performance. We needed to do precise weight tuning to avoid significant utility degradation while achieving substantial forgetting performance.

\vspace{1.0em}

We also found that LoRA gives considerable benefits for unlearning tasks in large pre-trained models such as CLIP. By constraining weight updates to low-rank matrices on projection layers instead of the full model weights, we obtain memory-efficient training while maintaining the ability to isolate and potentially reverse modifications. This architectural choice gave high value during our iterative experimentation phase as it enabled us to rapidly prototype different loss configurations without the computational costs of full model retraining. Moreover, LoRA's parameter efficiency allowed us to preserve the general capabilities of the base CLIP model while performing targeted adjustments on specific embeddings, resulting in minimal difference for preservation cases (absolute delta ranging from 0.0115 to 0.0608 across different preservation scenarios).

\vspace{1.0em}

Our thorough evaluation framework showed that effective safety unlearning requires extensive testing against diverse adversarial scenarios and utility preservation metrics. Indeed, the realization of paraphrase attacks, contextual attacks, and out-of-distribution image attacks brought about vulnerabilities that would not have been caught using only standard evaluation. The systematic measurement of utility preservation across single node preservation, new safe edges, new safe nodes, and neutral edges demonstrated that, although our model forgot the targeted unsafe relationships, general functionality was preserved. Such a multi-faceted evaluation approach—including both forgetting metrics and preservation metrics—established a rigorous validation standard for the effectiveness of unlearning. In this regard, going forward, safety-focused machine learning systems will clearly require not only sophisticated technical architectures but also comprehensive evaluation methodologies that can capture intended forgetting and unintended side effects.

\bibliographystyle{abbrvnat}
\bibliography{reference}

\end{document}